%% file: main.tex
\documentclass[journal]{IEEEtran}

\usepackage{amsmath,amssymb,amsfonts}
\usepackage{stmaryrd}
\usepackage{booktabs}
\usepackage{amsmath}
\usepackage{algorithm}
\usepackage{algpseudocode}
\usepackage{graphicx}
\usepackage{subfigure} 
\usepackage{float}
\usepackage{color}
\usepackage{comment}
\usepackage{caption}
\usepackage{soul,xcolor}
\usepackage{cite}
\usepackage{multirow}
\usepackage{colortbl}

\usepackage{hyperref}[colorlinks,
            linkcolor=red,
            anchorcolor=blue,
            citecolor=green
            ]

\makeatletter

\newcommand{\Rmnum}[1]{\expandafter\@slowromancap\romannumeral #1@}

\makeatother

\ifCLASSINFOpdf
\else
\fi

\begin{document}

\title{Probing the Robustness of Vision-Language Pretrained Models: A Multimodal Adversarial Attack Approach}
%  Multimodal Fusion Adversarial Attack for Vision-Language Pretrained Models
%Probing the Robustness of Vision-Language Pretrained Models: A Multimodal Adversarial Attack Approach

\author{Jiwei~Guan, Tianyu Ding, Longbing~Cao, Lei Pan, Chen~Wang, and Xi~Zheng
\thanks{Jiwei~Guan, Longbing Cao, and Xi~Zheng are with the School of Computing, Macquarie University, Sydney, Australia, e-mail: \{jiwei.guan, longbing.cao, james.zheng\}@mq.edu.au.}
\thanks{Tianyu Ding is with the Applied Sciences Group, Microsoft Corporation, Redmond, USA, e-mail: tianyuding@microsoft.com.}
\thanks{Lei~Pan is with the School of Information Technology, Deakin University, Waurn Ponds, Australia, e-mail: l.pan@deakin.edu.au.}
\thanks{Chen~Wang is with the Data61, CSIRO, Sydney, Australia, e-mail: chen.wang@data61.csiro.au.}
% <-this % stops a space}
}

% \markboth{IEEE Transactions on Neural Networks and Learning Systems}%
% {Guan \MakeLowercase{\textit{et al.}}: xxxxx}

% % \markboth{Preprint}%
% % {Guan \MakeLowercase{\textit{et al.}}: xxxxx}

\maketitle

\begin{abstract}
Vision-language pretraining (VLP) with transformers has demonstrated exceptional performance across numerous multimodal tasks. However, the adversarial robustness of these models has not been thoroughly investigated. Existing multimodal attack methods have largely overlooked cross-modal interactions between visual and textual modalities, particularly in the context of cross-attention mechanisms. In this paper, we study the adversarial vulnerability of recent VLP transformers and design a novel Joint Multimodal Transformer Feature Attack (JMTFA) that concurrently introduces adversarial perturbations in both visual and textual modalities under white-box settings. JMTFA strategically targets attention relevance scores to disrupt important features within each modality, generating adversarial samples by fusing perturbations and leading to erroneous model predictions. Experimental results indicate that the proposed approach achieves high attack success rates on vision-language understanding and reasoning downstream tasks compared to existing baselines. Notably, our findings reveal that the textual modality significantly influences the complex fusion processes within VLP transformers. Moreover, we observe no apparent relationship between model size and adversarial robustness under our proposed attacks. These insights emphasize a new dimension of adversarial robustness and underscore potential risks in the reliable deployment of multimodal AI systems.

\end{abstract}

% Note that keywords are not normally used for peer review papers.

\IEEEpeerreviewmaketitle

\input{sec1_introduction}

\input{sec2_relatedwork}

\input{sec3_methodology}
\input{sec4_experiment}

\input{sec5_ablation}
\input{sec6_conclusion}

\appendices

% \section*{Acknowledgement}
% We appreciate anonymous reviewers for their constructive feedback and valuable discussions. This work is in part supported by the CSIRO Data61 Collaborative Research Project (CRP) C020996, CSIRO Data61 Topup 194981059 and Australian Research Council Linkage Project (ARC) LP190100676. This work is also supported by the Taishan Scholars Program under Grant TSQN 202211214 and Shandong Excellent Young Scientists Fund Program (Overseas) No.~2023HWYQ-113.

\ifCLASSOPTIONcaptionsoff
  \newpage
\fi

\bibliographystyle{IEEEtran}
\bibliography{IEEEabrv,references}

\end{document}

%% file: sec1_introduction.tex
\section{Introduction}
\label{sec:introduction}

% 第一段介绍VLP 
Vision and Language Pretraining (VLP) employing transformers has recently emerged as a promising {approach} in understanding {the} complex relationships between visual and linguistic modalities~\cite{vlp2022du,chen2023vlp}. These VLP transformers enhance adaptability and acquire extensive knowledge by leveraging paired image and text inputs through self-attention learning. These models are pre-trained on large-scale datasets and subsequently fine-tuned  for a variety of specific downstream tasks~\cite{lu2019vilbert,li2020oscar,zhang2021vinvl}. Their performance has demonstrated great generalizability in a variety of multimodal vision and language tasks, including Visual Question Answering (VQA)~\cite{antol2015vqa,wu2017visual}, Visual Commonsense Reasoning (VCR)~\cite{zellers2019recognition} and Image Captioning~\cite{chen2015microsoft}.

% 第二段写危害 
% Despite the great success, VLP transformers exhibit a notable vulnerability to adversarial examples by subtle perturbation of the original data. This vulnerability persists even when pretraining models are fine-tuned for downstream tasks, potentially leading to erroneous outputs. More broadly, recent studies show common concerns about the vulnerability of multimodal generative AI systems like VLP transformers through adversarial attacks of deliberate manipulations~\cite{shayegani2023survey}. However, there has been limited research on the adversarial robustness of prevalent VLP transformers models without adversarial training, while they are increasingly used in security-sensitive real-world applications. To better understand the vulnerabilities of VLP transformers, it is essential to investigate how different modalities affect their performance under attack, as well as the adversarial robustness of recent VLP transformer cross-modal interactions. Furthermore, developing effective multimodal attack strategies for VLP transformers at the inference stage is crucial for evaluating their reliability in real-world scenarios. 

Despite the great success, VLP transformers exhibit a notable vulnerability to adversarial examples by subtle perturbation of the original data. This vulnerability persists even when pretraining models are fine-tuned for downstream tasks, potentially leading to erroneous outputs. More broadly, recent studies show common concerns about the vulnerability of multimodal transformer generative AI systems~\cite{shayegani2023survey}. However, there has been limited research to explore the adversarial robustness of prevalent VLP transformers without adversarial training. Moreover, the investigation of how cross-modal interaction affects adversarial attack performance has not yet been fully explored, especially when deployed in security-sensitive applications for commercial use.

% The following critical questions remain unanswered: 
% \begin{itemize}
%     \item[1)] How does attacking different modalities affect the model performance? 
%     \item[2)] What is the robustness of recent VLP transformer architectures against adversarial attacks? 
%     \item[3)] What constitutes an effective multimodal attack strategy for VLP transformers at the inference stage? 
% \end{itemize}

\begin{figure}[t]
\centering
\includegraphics[width=1.0\linewidth]{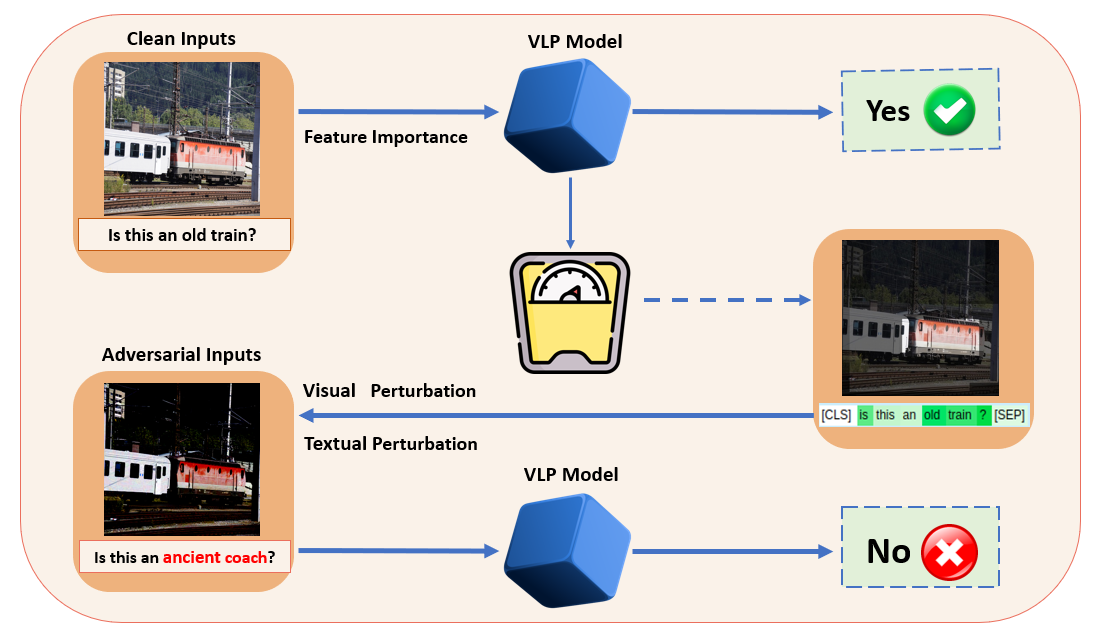}
\caption[]{JMTFA illustration of selecting trigger features for perturbing vision and language modalities on VisualBERT.} 
\label{fig:attnscore} 
\end{figure}

\begin{figure*}[ht]
\centering
\includegraphics[width=1.0\linewidth]{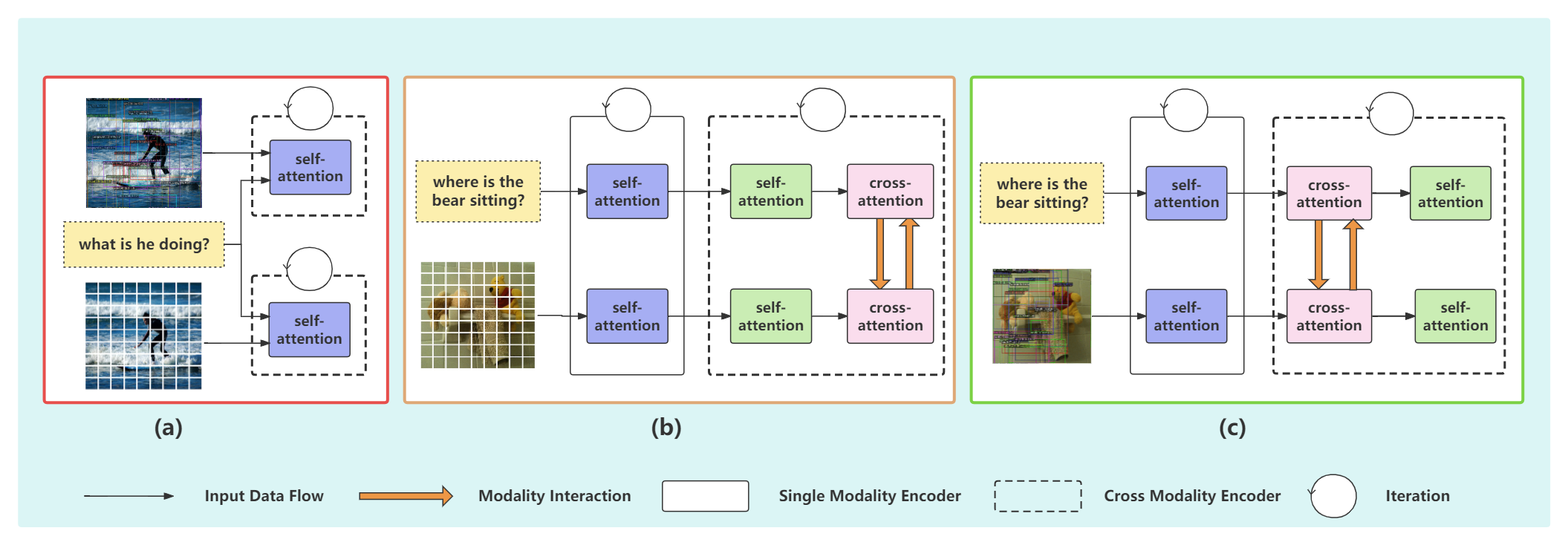}
\caption[]{Illustration of representative VLP transformer network architectures. (a) ViLT and VisualBERT: Single-stream architecture utilizing pure self-attention. (b) VLE: Dual-stream architecture combining self-attention with cross-attention. (c) LXMERT: Dual-stream architecture combining cross-attention with self-attention.}
\label{fig:modelarchitecture} 
\end{figure*}

% 第三段写贡献
% To address these questions, JMTFA adopts a white-box approach with modality fusion interactions to enhance the simultaneous perturbation of both vision and language modalities.

% to rigorously assess the sensitivity of victim VLP transformer models to adversarial perturbation,

% However, there is still a lack of complete understanding of how safety concerns arise from the fusion of vision-language modality.

% this paper aims to systematically analyze the adversarial robustness of VLP transformers. Thereby, 

Motivated by discovering cross-modality feature importance, this paper proposes a Joint Multimodal Transformer Feature Attack (JMTFA) approach that focuses on the intrinsic interactions between vision and language modalities, targeting the underlying factors that contribute to predictive outcomes. The proposed approach utilizes attention relevance scores derived from cross-modal engagement as a critical feature guide to simultaneously perturb both visual and textual modalities, as simultaneously optimizing the attack on both modalities rather than perturbing each modality separately has proven to be a more effective strategy~\cite{zhang2022towards}. This study discloses a deep understanding of the fusion processes in VLP transformers, integrating diverse and supplementary information to collaboratively boost multimodal adversarial attacks. As an illustration in Fig.~\ref{fig:attnscore}, JMTFA strongly shifts attention focal points that are the most relevant features of each modality, leading to misleading predictions. In detail, we primarily evaluate JMTFA on two typical fusion architectures of VLP by conducting experiments on vision-language understanding and reasoning benchmark tasks. These experiments conclusively demonstrate that JMTFA is highly effective in generating adversarial examples across both vision and language modalities. Furthermore, our findings reveal that there is no apparent relationship between adversarial resilience and the model size of the VLP transformers. Additionally, our experiments indicate that complex network architectures appear to rely more heavily on information from the textual modality than the visual modality. Our contributions are summarized as follows.
\begin{itemize}
    \item  A novel multimodal adversarial attack approach for VLP transformer models, which generates two-modality adversarial examples by leveraging the synergistic effect through cross-modality interaction fusion.
    \item  Systematical evaluation of the performance of adversarial attacks across various VLP models, revealing the VLP adversarial vulnerability in multiple downstream tasks.
    \item  Extensive experiments of perturbation optimization for attack effectiveness, which analyzes the impact of attack performance under various attack strategies 
\end{itemize}

%% file: sec2_relatedwork.tex
\section{Related Work} 
\label{sec:relatedwork}
% 相关工作1 VLP 单流双流  模型介绍 强调攻击的是fusion之后的模态表示 fusion会互相影响
\subsection{Vision and Language Pretraining}
VLP transformers typically comprise three essential components: a text encoder, a visual encoder, and a multimodal feature fusion encoder. The fusion encoder simultaneously acquires modality knowledge from text and visual encoders to effectively learn joint visual and textual feature representations, enhancing performance on multiple downstream tasks. In VLP fusion, there are two typical types of model architectures. Single-stream architectures, such as VILT~\cite{kim2021vilt} and visualBERT~\cite{li2020visualbert}, interact with modalities through self-attention, merging text and image features into a unified representation. In contrast, recent dual-stream architectures like VLE~\cite{iflytek2023vle} and LXMERT~\cite{tan2019lxmert} utilize separate transformers for encoding each modality. These dual-stream models employ cross-attention for more sophisticated information fusion, fostering richer modality-specific representations. As depicted in Fig.~\ref{fig:modelarchitecture}, these VLP architectures leverage advanced fusion in vision and language tasks. Despite impressive downstream task performances, there is a lack of studies on investigating adversarial vulnerability of fused VLP transformer models. The vulnerability of fusion module can provide an opportunity for adversaries who aim to deteriorate contextualize information  by exploiting modality attention influences.  In this research, our primary focus is on the adversarial robustness of discriminative VLP trasnformers in relation to classification challenges, leaving autoregressive generation VLP transformers like BLIP~\cite{li2022blip} and METER~\cite{dou2022empirical} for future exploration.

\subsection{Single Modality Adversarial Attack}

\textbf{Vision Modality Attacks}. Szegedy et al.~\cite{szegedy2014intriguing} first demonstrate adversarial vulnerability of deep neural networks where this study has been extensively studied for computer vision domains~\cite{goodfellow2015explaining,madry2018towards,papernot2016limitations,su2019one,moosavi2017universal,carlini2017towards}, focusing on optimizing the loss function with respect to single vision modality decision boundaries based on model outputs. Furthermore, feature importance attacks~\cite{wang2021feature,zhang2022enhancing} aggregate gradients concerning feature maps to perturb object-related regions. The feature disruptive attack~\cite{ganeshan2019fda} introduces an intermediate loss that alters the local activation of image features within the middle layer hidden outputs in target models, enhancing the transferability of perturbed features without requiring knowledge of their structures or parameters. \textbf{Language Modality Attacks}. Adversarial attacks have also emerged in natural language processing, impacting language model predictions through various manipulations at different levels. Character-level perturbations~\cite{gao2018black,eger2019text,boucher2022bad} involve character substitutions and visual character substitutions using swaps, deletions, insertions, and repetitions. These attacks utilize gradient-based and greedy-search methods. Word-level perturbations~\cite{ren2019generating,jin2020bert,garg2020bae} strive to replace key words with synonyms that have similar word and contextual embeddings. These methods use contextualized, optimization, greedy, and gradient-based word substitutions. Sentence-level perturbations~\cite{iyyer2018adversarial,ribeiro2018semantically,naik2018stress} generate paraphrases by applying rules, leveraging logical structures, or adding unrelated sentence segments. Despite extensive research on single-modality attacks, adapting these methods to VLP models remains largely unexplored.

\begin{figure*}[ht]
\centering
\includegraphics[width=\linewidth]{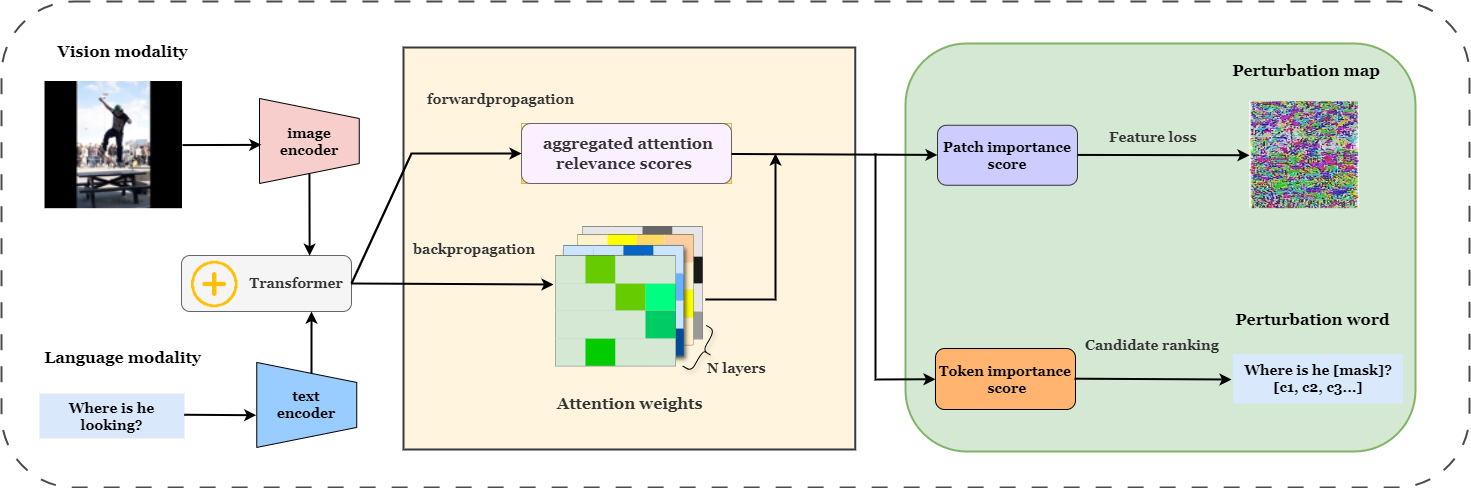}
\caption[]{An overview of JMTFA. The method uses cross-modal aggregated attention relevance scores to guide vulnerable features from vision and text modalities.}
\label{fig_attack_workflow} 
\end{figure*}

% 相关工作3 提出问题多模态对抗攻击只有检索
\subsection{Adversarial Attacks on VLP Models}
As VLP transformers proliferate, recent research has increasingly concentrated on multimodal adversarial robustness. In response to the vision and language vulnerability, Co-Attack~\cite{zhang2022towards} collectively executes attacks on both image and text modalities addresses potentially conflict by multiple single-source modality attacks to avoid in a counterproductive effect. VLATTACK~\cite{yin2022vlattack} organizes block-wise similarity to enlarge the distance between original and perturbed features of the image. If failed image attacking, this approach invokes BERT-Attack by employing the word-level replacement as a text attack. SGA~\cite{lu2023set} examines the transferability of adversarial examples in VLP transformers, utilizing modality interactions and alignments for information retrieval tasks. In addition, SA-Attack\cite{he2023sa} considers VLP transfer attacks incorporating inter-modal interaction and data diversity, using self-augmentation techniques 
to generate image and text adversarial examples. OT-Attack~\cite{han2023ot} applies optimal transport theory to determine the most efficient  correspondence between image and text features, guiding the creation of adversarial examples to improve adversarial transferability. Although these two-modality attacks highlight security requirements for the adversarial robustness of VLP transformers, they have overlooked the learning of mutual connections between fused intrinsic features of vision and language modalities. Furthermore, previous research has been validated on information retrieval tasks that utilize a single modality to search for top-ranked answers with other modality in a gallery database, which lacks generalizability on VLP understanding and reasoning tasks.

%% file: sec3_methodology.tex
\section{Methodology}
\label{sec:Methodology}
In this section, we introduce the JMTFA, a unified adversarial attack approach that simultaneously perturbs a pair of image and text. This approach details the adversary’s goals and capabilities to search feature importance through aggregated relevance attention scores. The importance of features as a key influential indicator in model predictions is then applied to generate adversarial visual and linguistic modality examples. As shown in Fig.~\ref{fig_attack_workflow}, JMTFA enforces forward propagation to produce a composite of relevance attention scores to pinpoint key features in the fusion of vision and language, where important features of one modality are influenced by the other modality in cross-modal interactions. A detailed description is provided below.

% 强调我们的方法带了梯度
\subsection{Recap Aggregated attention relevance scores}
We briefly describe the motivation for selecting aggregated relevance scores of attention in~\cite{chefer2021generic}. The attention-aggregated relevance scores involve the Hadamard products of attention scores and their gradients in each layer of multimodal transformers to characterize the influence of tokens on the prediction results.

\textbf{Single-Stream Architecture}. The aggregated attention relevance scores for a single-stream architecture are detailed in Equation~\ref{fma:selfrevelancemap}. In this context, $A$ represents the attention scores (product of queries matrix $Q$ and keys matrix $K$ within each attention layer), ${n}$ denotes the total amount of attention layers, ${i}$ represents the current layer, $\nabla A$ refers to the backpropagated gradients from the outputs to the attention layer, and $\odot$ denotes element-wise multiplication. The result of the element-wise multiplication is an attention-weight matrix, which is normalized as described in~\cite{bach2015pixel} and considers only the positive contributions as outlined in~\cite{abnar2020quantifying}. In addition, ${E}_h$ represents the mean head values (average of positive normalized attention weight matrix across all heads for each attention layer), and $\hat{A}$ is the aggregated attention relevance score. With such accumulation, the aggregated attention relevance scores by a forward pass contribute to the feature importance. 
\begin{equation}
\hat{A} = \prod \limits_{i=1}^n \mathbb{E}_h\left[( A \odot \nabla A)^+\right]_{\text{norm}} 
\label{fma:selfrevelancemap}
\end{equation}

\textbf{Dual-Stream Architecture}. Dual-stream architectures incorporating both self-attention and cross-attention layers, are designed to extract meaningful image and text representation features by an independent transformer. This architecture enables each modality to be influenced by the other through preceding attention layers via matrix multiplication, and vice versa. Additionally, each modality retains influence from itself in both previous and current layers. It computes aggregated attention relevance scores that contextualize the vision and language modalities, which can be formalized as follows:
\begin{equation}
\hat{A} = {A_s} + \prod \limits_{i=1}^n \mathbb{E}_h\left[( A_c \odot \nabla A_c)^c\right]_{\text{norm}}
\label{fma:crossrelevancemap}
\end{equation}
where ${A_s}$ represents the aggregated self-attention relevance scores for each modality, as detailed in Equation~\ref{fma:crossrelevancemap}, $A_c$ is defined as the cross-attention scores derived from the cross-attention weights of the vision or language modality in each layer, $\nabla A_c$ represents the gradients obtained through backpropagation, and the superscript $^c$ indicates the contextually mutual interactive relevance from the other modality. Overall, $\hat{A}$ combines the aggregated self-attention relevance scores $A_s$ with the aggregated cross-attention relevance scores $A_c$, providing a comprehensive representation of the attention dynamics in the dual-stream architecture.

\subsection{Vision Modality Attack}
For the JMTFA on the image modality, we utilize aggregated attention relevance scores to highlight object-related regions and mitigate interest in non-object regions. {This approach of identifying feature importance through aggregated attention relevance scores motivates us to preserve the disrupted direction towards adversarial examples. Given this, we define the loss function of the Attention Feature Importance Attack (AFIA) as follows:
\begin{equation}
\mathcal{L}\left(x_i^{\text{adv}}\right) = \mathcal{L}_{\text{AFIA}}\left(  \hat{A}\odot f_{vlp}\left(x_i, x_t\right) \right)
\label{fma:fia}
\end{equation}
where $\hat{A}$ represents the image aggregated attention relevance scores that explicitly indicate important image features (patches), $x_i$ and $x_t$ denote the image and text inputs, respectively, and $f_{vlp}$ is the VLP model for image features with respect to a true label. 

This proposed AFIA loss function aims to decrease the important feature with positive contributions to the vision modality. In image attack, the practical generation of adversarial image examples involves two steps: 1) Constructing the loss function optimization, where the optimization procedure employs an iterative process to escape local optima, and} 2) Restricting and implementing noise, where the adversary designs practical schemes to apply attacking perturbations under perceivability constraints. The Equation~\ref{fma:pgdjmtfa} illustrates the PGD iteration for generating adversarial examples for images:
\begin{equation}
\begin{aligned}
& \quad x_{i}' = \text{Clip}_{x_t, \epsilon} \left\{ x_i + \alpha \cdot \text{sign}\left(\nabla_{x_i} \mathcal{L}_{\text{AFIA}}(x_i, x_t)\right) \right\}
\end{aligned}
\label{fma:pgdjmtfa}
\end{equation}
where $\epsilon$ is the maximum perturbation and $\alpha$ is the step size. This iterative process allows for the generation of adversarial image examples while maintaining perceptual constraints on the vision modality.

\subsection{Language Modality Attack} 
For the JMTFA on the language modality, we adopt BERT-Attack~\cite{li2020bert} to determine word importance using text aggregated attention relevance scores $\hat{A}$, rather than relying solely on logits from the victim models. Although gradient-based attacks are efficient for images, they are challenging to implement for textual attacks due to naturalness constraints, including lexical, semantic, and grammatical considerations. BERT-Attack strategically applies the masked language model to generate various adversarial substitute candidate tokens, excelling in strong word replacement using contextual perturbations while retaining sentence semantics. Moreover, word-level attacks are more effective than sentence-level attacks and stealthier than character-level attacks~\cite{zhang2023towards}. 

The text attack of JMTFA comprises two stages: 1) Identifying important words using language modality aggregated attention relevance scores to maximize the chance of incorrect predictions by the victim model, and 2) ensuring semantic integrity by generating top-$K$ candidate tokens that strive to maintain semantic continuity and ensure minimal perturbation levels. The following Equation~\ref{fma:bertattackjmtfa} demonstrates the JMTFA for generating text adversarial examples:
\begin{equation}
x_t' = \arg \max_{x_t' \in \mathcal{X}} P(y' \mid x_t') \quad \text{s.t.} \quad \|x_t - x_t'\|_0 \leq \epsilon, \; \epsilon \in \hat{A}
\label{fma:bertattackjmtfa}
\end{equation}
where $x_t'$ represents the adversarial text example, $x_t$ is the original text input, and $\epsilon$ is the maximum perturbation, constrained to $K$ tokens. This allows for the generation of adversarial text examples while maintaining semantic consistency and minimizing the number of modified tokens, as guided by the aggregated attention relevance scores.

% Language modality attack first detects whether candidate perturbed samples can change the prediction or reduce true output probability.
% 可能没有合适的词汇替换所以同时攻击

\subsection{Vision and Language Modality Attack} 
In a white-box scenario, the adversary has complete knowledge of architecture and parameters of the victim VLP models, enabling the identification of aggregated relevance attention scores as feature importance. Perturbing the two modalities by combining Formulations in~\ref{fma:pgdjmtfa} and \ref{fma:bertattackjmtfa} strives to find vulnerabilities across the fused intrinsic feature space. BERT-Attack may not generate appropriate word substitutions or can change the prediction. To overcome this, JMTFA simultaneously targets the vision modality and dynamically update perturbations under the guidance of cross-modal interactions. The overall scheme of JMTFA in an adversarial manner is summarized in Algorithm~\ref{alg:FAARMA}. JMTFA formulates the problem of generating adversarial examples based on either single modality or two modalities. This approach allows for leveraging the interdependencies between vision and language modalities, leading to exploit the vulnerabilities in cross-modal interactions. Consequently, the proposed JMTFA can be used in a more variety of vision and language transformer architectures due to the attention mechanism to learn visual and linguistic features.

% Perturbation two-modality by combining Formulations in~\ref{fma:pgdjmtfa} and \ref{fma:bertattackjmtfa} strives to maintain the naturalness of two-modality inputs to make the perturbation undetectable to humans.

\begin{algorithm}[ht]
\small
\caption{Joint Multimodal Transformer Feature Attack}
\label{alg:FAARMA}
\begin{algorithmic}[1] % The number enables line numbering
\State \textbf{Input:} Original text $t$ and image $i$, multimodal transformer classifier image feature $f()$, relevance scores $\hat{A}$.
\State \textbf{Output:} Adversarial image $i^{\text{adv}}$ and text $t^{\text{adv}}$
\State \textbf{Initialize:} Step size $\alpha$, max perturbation $\epsilon$, iterations $N$, candidate number $K$.
\State \textbf{Modality attention relevance scores $\hat{A}_i, \hat{A}_t \gets \hat{A}$}

\State \textbf{Update $i^{\text{adv}}$ by feature importance method:}

\State {Construct loss optimization objective:}
\State \hspace{1em} $\mathcal{L}(i^{\text{adv}}) = \mathcal{L}(\hat{A_i}\odot f(i, t))$

\Procedure{Feature importance ranking $\hat{A_i}$}{}
\For{$n = 0$ to $N-1$}
    \State $g = \nabla_i \mathcal{L}(i^{\text{adv}})$
    \State $i_{n+1}^{\text{adv}} = \text{Proj}_{i, \epsilon}\{i_{n}^{\text{adv}} + \alpha \cdot \text{sign}(g)\}$
\EndFor
\EndProcedure

\State \textbf{Guide $t^{\text{adv}}$ BERT attack method:}
\Procedure{Word importance ranking $\hat{A_t}$}{}
\State Choose a word list $L \gets [w_{\text{top-1}}, w_{\text{top-2}}, \ldots]$ in descending order by $\hat{A_t}$
\State Search top-$K$ candidate list $C \gets [c_0, \ldots, c_{j}]$ using BERT for $P^\in \mathbb{R}^{n \times K}$
\For{$w_j$ in $L$}
    \State Get candidate adversarial word $C \gets \text{Filter}(P_j)$
    \State Change selected word $w_{j}$ by $t^{\text{adv}} \gets [w_0, \ldots, w_{j-1}, c_k, \ldots]$ 
\EndFor
\EndProcedure

\State \Return $i^{\text{adv}}$, $t^{\text{adv}}$
\end{algorithmic}
\end{algorithm}

%% file: sec4_experiment.tex
\section{Experiments}
\label{sec:Experiment}

\subsection{Models and Datasets} 
% 介绍受害模型
% Recent advancements in vision and language tasks have increasingly relied on attention mechanisms and feature combination strategies to significantly enhance the accuracy of predicted answers. 

In our experiments, we primarily conduct experiments on these four VLP models in our paper: ViLT~\cite{kim2021vilt} and VisualBERT~\cite{li2020visualbert}, representing single-stream architecture, and VLE~\cite{iflytek2023vle} and LXMERT~\cite{tan2019lxmert}, exemplifying dual-stream architecture. Single-stream architectures integrate visual and textual embeddings into a unified sequence, utilizing self-attention for modality fusion. In contrast, dual-stream architectures use both cross-attention and self-attention layers to facilitate cross-modal fusion. These models differ in their image and text encoding methods. Specifically, ViLT employs the pixel-level visual transformer (ViT)~\cite{dosovitskiy2021an} for its image encoding, while VLE utilizes CLIP-ViT\cite{iflytek2023vle}. On the other hand, VisualBERT and LXMERT rely on the regional-level pre-trained object detector Faster-RCNN~\cite{ren2016faster}. For textual encoding, ViLT, VisualBERT, and LXMERT implement BERT~\cite{kenton2019bert}, whereas VLE utilizes DeBERTa-v2~\cite{he2022debertav3}. All selected VLP models process image-text pairs as inputs, with the primary distinction lying in their methods of encoding visual features. We use these VLPs from the Huggingface hub to evaluate the JMTFA on different architectures, focusing on fine-tuned public VLP models rather than direct inferences from off-the-shelf pretraining models due to their relatively lower classification performance.

% 介绍数据集 vqa 5000是要覆盖住3129个答案标签 VSR使用1000
Our evaluation focuses on two critical tasks: \textbf{Visual Question Answering} (VQA) refers to vision and language understanding  and \textbf{Visual Spatial Reasoning}~\cite{liu2022vsr} (VSR) indicates vision and language reasoning. Understanding and reasoning capabilities are pivotal for multimodal AI systems that require deeper, sophisticated interactions between visual and linguistic domains, such as in-field robotics and autonomous driving. The VQA task requires accurately comprehending natural questions and real-world images to respond with precise answers, formulated as a classification problem with 3,129 predefined most frequent responses. Images in VQAv2 are from Microsoft COCO~\cite{lin2014microsoft} and the dataset contains 65 types of open-ended questions, with answers classified into three types: yes/no, number, and other. Similarly, the VSR task predicts locations based on image details under textual template instructions, capturing complex relational information within an environment. This task requires recognizing spatial relations among objects in a manner akin to human cognition, articulated through natural language that encompasses intricate lexical and syntactic details. We experiment on the VQAv2 validation set using 5,000 natural language questions with associated images and ground truth answers. The VSR evaluation is performed on a test dataset comprising 1,000 image-text pairs and uses a zero-shot model that has no overlapping training location concepts. It is worth noting that no VSR fine-tuned model exists for VLE.

\subsection{Implementation Details and Metrics} 

 % a controlled dataset that explicitly tests VLMs for spatial reasoning
To instantiate image attack, we resize the images to $384 \times 384$, except for VLE that uses $576 \times 576$ due to its default input settings. We adopt the PGD optimizer with a budget $\epsilon$ of 4/255 or 8/255, a step size $\alpha$ of 2/255, and a fixed total iteration number of 40. All perturbations are constrained within a bound of 8/255 under the \( \ell_{\infty} \) norm, except for VLE, where $\epsilon$ is set to 14/255 based on empirical observations. For the text attack, we utilize the default settings of BERT-Attack. We adhere to the default number of candidate words as 48 and use the similar-words-filter matrix~\cite{jin2020bert} for antonyms. The maximum perturbation $\epsilon$ is set to $K=3$ tokens. We investigate both single-source modality and dual-source modality attacks on each iteration to explore  worst-case permutations.

We employ two primary metrics to evaluate JMTFA on victim VLP models: i) Attack Success Rate (ASR), which is the proportion of adversarial examples that effectively fool target VLP models among all correct predictions; and ii) Adversarial Accuracy, which is the classification rate on the whole testing dataset under adversarial attacks. We report the best attack performance achieved by adding perturbations to the legitimate image and text pair. These metrics effectively evaluate two-modality adversarial examples and their impact on the performance of the target VLP models.

\begin{table*}[t]
\small
    \centering
    \setlength{\tabcolsep}{5pt} % 调整列间距
    \renewcommand{\arraystretch}{1.2} % 调整行间距
    \begin{tabular}{l|rr|rrr|c|c|c}
        \hline
        \textbf{Model} & \multicolumn{8}{c}{\textbf{VQAv2 (5K val set)}} \\
        \cline{2-9}
         & \multicolumn{2}{c|}{\textbf{Vision Only}} & \multicolumn{3}{c|}{\textbf{Language Only}} & \multicolumn{2}{c|}{\textbf{Multimodality}} & \textbf{Transferable} \\
        \cline{2-9}
         & $\epsilon$@4/255 & $\epsilon$@8/255 & $k$@1 & $k$@2 & $k$@3 & $\epsilon$@4/255 \& $k$@3 & $\epsilon$@8/255 \& $k$@3 & $\epsilon$@8/255 \& $k$@3 \\
        \hline
        ViLT & 55.75 & 57.98 & 18.69 & 34.11 & 50.95 & 81.48 & 85.52 & - \\
        \hline
        VisualBERT & 52.93 & 53.42 & 28.41 & 32.74 & 37.68 & 76.07 & 76.49 & 80.76 \\
        \hline
        LXMERT & 52.42 & 55.50 & 33.93 & 48.69 & 59.45 & 72.72 & 84.14 & 82.62 \\
        \hline
        VLE$^\dagger$ & 52.83 & 68.06 & 26.54 & 51.03 & 66.49 & 80.50 & 88.27 & -\\
        \hline
    \end{tabular}
    \caption{Attack performance on the VQAv2 dataset, evaluated across single and dual modalities. Results are reported as ASR (\%), with higher values indicating more effective attacks. In VLE, $^\dagger$ symbol indicates $\epsilon$ set to 8/255 and 14/255.}
    \label{tab:vqa_asr}
\end{table*}

\begin{table*}[hbtp]
    \small
        \centering
        \begin{tabular}{|l|c|c|c|c|c|c|c|c|}
        \hline
        \hline
        \textbf{Model type} &  \textbf{Number}   & \textbf{Color}  &  \textbf{Yes/No}  & \textbf{Others} & \textbf{Overall}  & \textbf{descent in ACC}  \\
        \hline
        ViLT &  2.10\% & 1.30\% & 28.30\% & 3.40\% & 13.10\% & 57.23\%   \\ 
        \hline
        VisualBERT & 1.30\%  & 3.50\% &  40.10\%  & 0.70\% &  16.64\% & 54.16\%  \\ 
        \hline
        VLE &  2.51\%&  0.00\%  & 23.00\% & 1.8\% & 9.7\%  &  67.90\%     \\
        \hline
        LXMERT & 5.50\% &  1.40\% &  3.82\% & 8.10\% & 12.60\% &  59.90\%  \\
        \hline
        \end{tabular}
        \caption{Evaluation of attack accuracy for various question types under two-modality JMTFA}
    \label{tab:jmtfa_variousacc}
\end{table*}

\subsection{Attack Performance on VQA}
The results of JMTFA on the VQA tasks are shown in Table~\ref{tab:vqa_asr}. We notice that LXMERT achieves the highest ASR of 68.06\% in the JMTFA image attack, while VLE has the highest ASR of 80.50\% in the JMTFA text attack. Additionally, we used ViLT and VLE as source models to investigate the JMTFA transferability for VisualBERT and LXMERT separately. Experimental results demonstrate pixel-level perturbations are not effective as object level for dual-architecture, but stronger than single-architectures, denoting that the transferability is inconsistent. Overall, JMTFA achieves the best ASR of 88.27\% in VLE. Our analysis reveals several key insights: (1) Object detection significantly enhances the intrinsic feature representation of objects. Improving regional features of the visual modality can achieve strong competitive adversarial robustness. (2) Perturbing two-source modality are more effective than perturbing any single-source modality (image or text), demonstrating that stronger adversarial attacks are required when multiple modalities are involved.

(3) Dual-stream architecture models seem to rely more on textual information than single-stream architecture models, as evidenced by the JMTFA algorithm's alternative perturbations to the most salient words. (4) Attack performance in different VLP models suggests that there is no significant impact on adversarial robustness between single-stream and dual-stream architectures.

\begin{figure}[t]
\centering
\includegraphics[width=.85\linewidth]{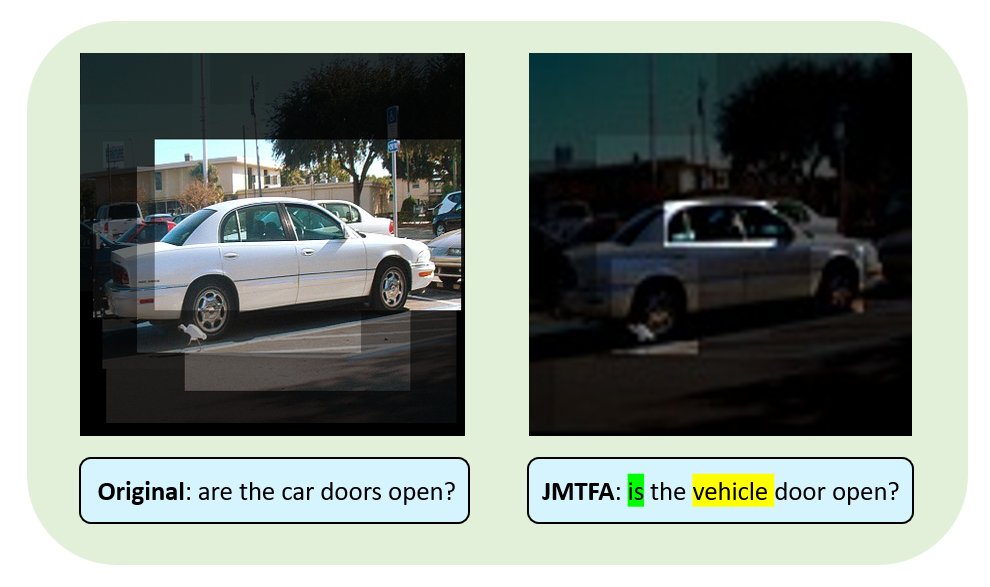}
\caption[]{JMTFA shifts attention away from ``doors" while simultaneously altering associated visual segments with text guidance. Color options such as green and yellow indicate perturbed words in the benign text.} 
\label{fig:attnvis} 
\end{figure}

\noindent\textbf{VQA Adversarial Example Visualization}. Fig.~\ref{fig:attnvis} illustrates the image and text attention visualization before and after JMTFA. The left column shows attention interest areas of clean inputs, while the right column displays adversarial examples under two modality attacks. JMTFA can redirect attention from ``doors'' to different visual segments of the object car. Our findings reveal that JMTFA causes a dramatic shift in attention focus toward the perturbed components producing incorrect answers contrary to the expected VLP model behavior. The right column demonstrates that adding two-modality adversarial perturbations by JMTFA significantly influences risk-sensitive decisions related to important features in the specific domain.  These results underscore the security vulnerabilities of VLP models and highlight the effectiveness of JMTFA in exploiting these weaknesses.

\begin{table*}[t]
\small
\centering
\setlength{\tabcolsep}{4pt} % 调整列间距
\renewcommand{\arraystretch}{1.2} % 调整行间距
\begin{tabular}{l|cc|ccc|c|c|c|c}  

\hline
\textbf{Model} & \multicolumn{9}{c}{\textbf{VSR (1K test set)}} \\
\cline{2-10}
 & \multicolumn{2}{c|}{\textbf{Vision Only}} & \multicolumn{3}{c|}{\textbf{Language Only}} & \multicolumn{2}{c|}{\textbf{Multimodality}} & \multicolumn{2}{c}{\textbf{Multimodality-MSE}} \\
\cline{2-10}
 & $\epsilon$@4/255 & $\epsilon$@8/255 & $k$@1 & $k$@2 & $k$@3 & $\epsilon$@4/255 \& $k$@3  & $\epsilon$@8/255 \& $k$@3 & $\epsilon$@4/255 \& $k$@3 & $\epsilon$@8/255 \& $k$@3 \\
\hline
ViLT & 23.66 & 38.37 & 15.90 & 26.71 & 36.77 & 51.50 & 59.30 & 76.72 & 91.90 \\
\hline
VisualBERT & 25.42 & 26.86 & 29.86 & 30.58 & 32.00 & 51.30 & 53.00 & 69.00 & 86.29 \\
\hline
LXMERT & 24.52 & 37.20 & 25.48 & 44.11 & 59.73 & 53.60 & 53.97 & 89.42 & 91.20 \\
\hline
\end{tabular}
\caption{Attack Performance on VSR Dataset. The reported value is ASR (\%).}
\label{tab:vsr_asr}
\end{table*}

\noindent\textbf{Analysis Question Types in VQA}. We further examine which types of questions are most vulnerable to JMTFA. We maintain default settings for fairness and use regional features due to their attack performance across all VLP models. Table~\ref{tab:jmtfa_variousacc} presents a breakdown of attack performance across different question types. For each question type, perturbed features in vision and language modalities decrease performance to varying degrees on different VLP models. We observe that the Yes/No category is the most challenging for the model to fool, while the Color category is the easiest to attack. Notably, we found no clear correlation between robustness and model size under JMTFA attacks.

% JMTFA struggles with adversarial examples in the perturbation space for short, simple, and well-represented questions.

% 最强的攻击会改变语义 是否写入文章  Using default settings in the language modality

\noindent\textbf{Human Evaluation}. We also perform a qualitative human evaluation on the adversarial texts. This evaluation aims to determine human judgment rates for semantic consistency and grammatical correctness. We recruit five undergraduate student volunteers who are native English speakers to participate in the study. These evaluators are presented with 1,000 pairs of original and adversarial questions and asked to score the semantics and grammar on a scale of 0-1. Table~\ref{tab:human_evaluation} presents the semantics and grammar scores from human evaluations compared to automated sentence semantic similarity measurements~\cite{reimers2019sentence}. The results indicate that adversarial questions generated by JMTFA with worst perturbations are largely indistinguishable from original questions in human evaluations. This demonstrates the effectiveness of JMTFA in creating adversarial examples that maintain semantic and grammatical coherence while successfully fooling VLP models. However, we also identify a main drawback: the occasional generation of out-of-domain words from human perspectives. This finding highlights the need for further refinement in maintaining domain-specific vocabulary in generated adversarial questions.

\begin{table}[htbp]
\small
\centering
\begin{tabular}{cccc}
\hline\hline
\textbf{Model} & \textbf{Similarity} & \textbf{Semantic} & \textbf{Grammar} \\ \hline
ViLT           & 0.55                        & 0.60              & 0.45             \\ 
VisualBERT     & 0.52                        & 0.44              & 0.56             \\ 
VLE            & 0.54                        & 0.60              & 0.40             \\ 
LXMERT         & 0.66                        & 0.42              & 0.49             \\ 
\hline\hline
\end{tabular}
\caption{Comparison of human evaluation results with semantic textual similarity scores.}
\label{tab:human_evaluation}
\end{table}

\subsection{Attack Performance on VSR}
% 两个实验 写改变了损失函数 另外vilt lxmert有位置信息 容易被攻击
% 1 双模态效果好 2 加入标签信息更好 3双流更鲁邦 4 mse是拉进距离 效果差不多 5 visualbert jmtfa效果好 是100个 objects
% We report the attack experimental results on the VSR task in Table~\ref{tab:vsr_asr}. Concretely, our primary findings are as follows: (1) Perturbing two modalities outperforms perturbing a single source modality, which is consistent with the VQA challenge. (2) Adopting the language modality of JMTFA has revealed stronger attacks than the vision modality of JMTFA on VLP models, indicating less impact on the image compared to the text. (3) LXMERT has shown stronger adversarial robustness than ViLT and VisualBERT in vision attacks by JMTFA. (4) VisualBERT has higher ASR values in VSR than VQA due to 100 objects by default settings. (4) JMTFA on VisualBERT outperforms ViLT and LXMERT, highlighting the importance of positional encodings of VLPs, as ViLT uses patches by positional encodings and LXMERT has object regions with position features. (5) Consequently, the impact of attacking the vision modality is relatively weak for VSR. Regarding these observations, JMTFA can add the logits and label as pair functions to FIA in Equation~\ref{fma:fia}, denoted as JMTFA-MSE. JMTFA-MSE to minimum FIA and semantic labels distance proves to be more effective than FIA. 

% TODO 写出原因 虽然能改变注意力关注的区域 但是不能判断位置
We report the attack experimental results on the VSR task in Table~\ref{tab:vsr_asr}. Our observations indicate that the impact of attacking the vision modality is relatively weak for VSR. To address this, we introduce JMTFA-MSE, which adds a loss distance function to AFIA in Equation~\ref{fma:fia}. JMTFA-MSE, designed to measure feature importance and semantic label distance when the vision modality attack, proves more effective than FIA alone. Our findings are as follows:  (1) Perturbing two modalities consistently outperforms perturbing a single source modality, aligning with our observations from the VQA challenge. (2) The language modality of JMTFA demonstrates stronger attacks than the vision modality on VLP models, suggesting that text perturbations have a more significant impact. (3) VisualBERT exhibits stronger adversarial robustness than ViLT and LXMERT, highlighting the importance of positional encodings in VLPs. This is evident in ViLT's use of patches for positional encodings and LXMERT's incorporation of object regions with position features.

% \begin{itemize}
%     \item Perturbing two modalities consistently outperforms perturbing a single source modality, aligning with our observations from the VQA challenge.
%     \item The language modality of JMTFA demonstrates stronger attacks than the vision modality on VLP models, suggesting that text perturbations have a more significant impact than image perturbations.
%     % VisualBERT shows higher ASR values in VSR compared to other models, which may be attributed to the default setting of 100 objects
%     % MTFA's performance on VisualBERT surpasses that of ViLT and LXMERT,
%     \item VisualBERT exhibits stronger adversarial robustness than ViLT and LXMERT, highlighting the importance of positional encodings in VLPs. This is evident in ViLT's use of patches for positional encodings and LXMERT's incorporation of object regions with position features.
% \end{itemize}

\begin{figure}[htbp]
\centering
\includegraphics[width=1.0\linewidth]{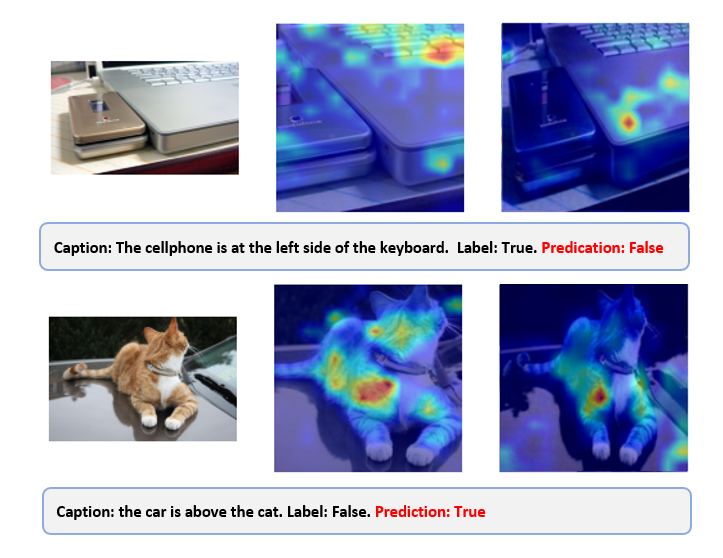}
\caption{
A comparison of attention maps for benign images in ViLT versus adversarial attention maps in VSR, demonstrating the shift in attention patterns under the proposed attack.
% ViLT benign image examples with attention maps compared to adversarial attention maps in VSR, showing attention shifting under the proposed attack.
}
\label{fig:attckheatmap}
\end{figure}

\noindent\textbf{VSR Adversarial Example Visualization}. Fig.~\ref{fig:attckheatmap} illustrates the effectiveness of our attack method. Each row contains the original image, alongside original and adversarial attention interpretability maps. The adversarial examples demonstrate how ViLT fails to capture important objects and recognize their orientation. In the examples, the victim ViLT model struggles with understanding left-right relations such as ``at the right of" or ``the left side of" a cellphone and a keyboard, respectively, resulting in incorrect predictions. The second row consistently shows misunderstanding of the ``above" relation between objects. JMTFA's success can be attributed to its focus on the final attention layers, which are more semantic in nature. This approach helps promote trivial features that effectively mislead victim VLP models, as clearly and consistently visualized by the attention maps.

\begin{table}[t]
    \centering
    \resizebox{\columnwidth}{!}{%
    \begin{tabular}{c|ccc|c}
    \hline\hline
    \textbf{Location} & \textbf{ViLT} & \textbf{VisualBERT} & \textbf{LXMERT} & \textbf{Frequency} \\ \hline
    under        & 0.4891 & 0.6520 & 0.3587 & 92 \\ 
    contains     & 0.3077 & 0.3205 & 0.1667 & 78 \\ 
    touching     & 0.3784 & 0.1081 & 0.2838 & 74 \\
    on top of    & 0.3803 & 0.2820 & 0.3099 & 71 \\
    on           & 0.4928 & 0.0435 & 0.3478 & 69 \\
    in           & 0.5238 & 0.0000 & 0.3968 & 63 \\
    inside       & 0.4681 & 0.1489 & 0.2766 & 47 \\ 
    beneath      & 0.4130 & 0.2609 & 0.2391 & 46 \\
    surrounding  & 0.2581 & 0.1290 & 0.2258 & 31 \\
    in front of  & 0.2759 & 0.0690 & 0.4138 & 29 \\
    \hline\hline
    \end{tabular}%
    }
    \caption{Prediction performance of relations under JMTFA.}
    \label{tab:vsrlocations}
\end{table}

% 放三个不同的图 比较direction不同 写出改变答案内容 看出答案更偏向哪里
% we find the last the layer, the better it performs, so we choose the last layer as default 

\begin{table*}[ht]
\small
\centering
\begin{tabular}{c|cccccccc}
\hline\hline
\textbf{Model} & \textbf{Task} & \textbf{FGSM + PWWS} & \textbf{PGD + PWWS} & \textbf{FGSM + BA} & \textbf{PGD + BA} & \textbf{COA} & \textbf{JMTFA}  & \textbf{JMTFA-MSE}  \\
\hline
\multirow{2}{*}{\textbf{ViLT}} & VQA & 65.59 & 80.69 & 69.94 & 78.44 & 51.05 & \cellcolor{gray!30} 85.52  & \cellcolor{gray!30}  89.02 \\
 & VSR & 34.73 & 32.82 & 31.81 & 31.42 & 36.39 & \cellcolor{gray!30}  59.30  &  \cellcolor{gray!30}  91.90\\
\hline
\multirow{2}{*}{\textbf{VisualBERT}} & VQA &  66.72  & 67.46 & 67.51 & 73.10 & 69.60 & \cellcolor{gray!30}  76.49   &  70.19\cellcolor{gray!30} \\
 & VSR & 17.71 & 25.42 & 25.28 & 34.86 & 27.14 & \cellcolor{gray!30}  53.00 & \cellcolor{gray!30}  90.40\\
\hline
\textbf{VLE} & VQA & 76.26 & 80.70  & 69.03 & 64.43& 68.29 & \cellcolor{gray!30}  88.27 & \cellcolor{gray!30}  95.50  \\
\hline
\multirow{2}{*}{\textbf{LXMERT}} & VQA & 57.54 &  69.05 & 70.84 & 94.34 & 69.79  & \cellcolor{gray!30}  80.20 & \cellcolor{gray!30} 94.48\\
 & VSR & 39.63 & 36.30  & 42.55 & 50.36 & 39.18 & \cellcolor{gray!30} 53.97 & \cellcolor{gray!30} 91.20 \\
\hline\hline
\end{tabular}
\caption{Comparison of JMTFA with baselines results on VLP models for VQA and VSR tasks as reported by ASR (\%). BA represents BERT-Attack and COA is CO-Attack.}
\label{tab:jftma_baseline}
\end{table*}

% We conjecture this is due to the fact that APGD-S is too effective

\begin{figure}[htbp]
\centering
\includegraphics[width=.95\linewidth]{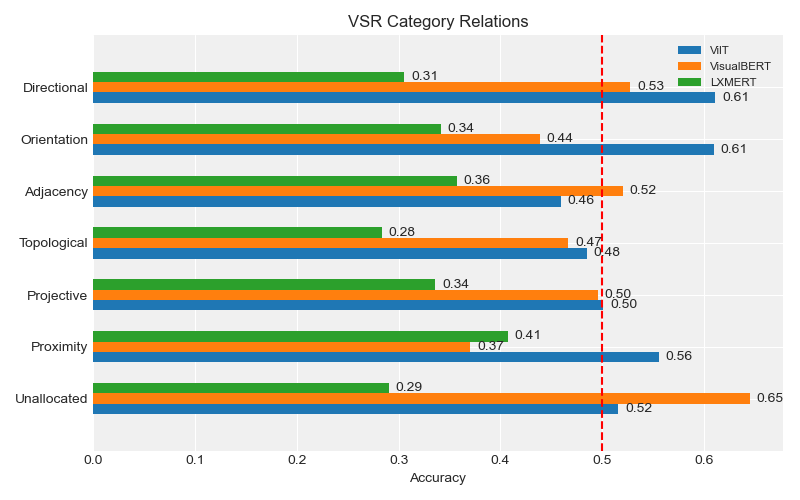}
\caption[]{} 
\label{fig:vsrrelations}
\caption{Performance of JMTFA across different categories of relations.}
\end{figure}

\noindent\textbf{Analysis Relation Types in VSR}. Table~\ref{tab:vsrlocations} presents attacking predictions for spatial locations and accuracy, highlighting diverse results for the top 10 frequency items. These linguistic observation fluctuations illustrate no preference for predicated locations, and attack accuracy varies across different VLP models. For example, the ``in'' predicting location accuracy shows variability, {suggesting no specific spatial language correlation between various VLP models. We further categorize spatial relations into seven groups as defined in~\cite{marchi2021cross}: ``Adjacency'', ``Directional'', ``Orientation'', ``Projective'', ``Proximity'', ``Topological''  and ``Unallocated''. As shown in Fig.~\ref{fig:vsrrelations}, JMTFA leads to a moderate decline in accuracy performance across all victim VLP models and categories. Compared to the original paper, accuracy decreases to almost under 50\% across all models, with ViLT showing a particularly drastic decline. Interestingly, VisualBERT maintains around 65\% accuracy in the ``Unallocated'' category}, indicating its adversarial robustness without spatial features reasoning. The ``Directional'' category demonstrates the lowest classification performance on the test dataset, with attack accuracy across all fine-tuned VLP models in this category averaging below chance level.

 % These questions are typically querying the peripheral aspects of the image instead of the main object, and actually require more complex reasoning
% 是否作为消融实验 因为VSR还有random 模型

% 可以换成表格展示图太占空间, 
% \begin{figure}[htbp]
% \centering
% \includegraphics[width=1.0\linewidth]{figures/vsr_direction.png}
% % \includegraphics[width=9cm,height=5.5cm]{figures/}
% \caption[]{Prediction performance of relations under JMTFA} 
% \label{fig:attackdirection} 
% \end{figure}

% To compare with gradient-based combination attacks, we incorporate \textit{FGSM+PWW}, \textit{PGD+PWWS}, \textit{FGSM+BERT-Attack}, \textit{PGD+BERT-Attack}, \textit{Co-Attack}. Table~\ref{tab:jftma_baseline} displays the comparative findings for the VQA and VSR task, revealing the following insights: (1) For a fair comparison, JMTFA employs the strongest attack settings by constraints, JMTFA outperforms superior performance over combined multimodal adversarial attack and Co-Attack baselines. This substantiates the reliability of the analysis results. (2) JMTFA surpasses baseline attacks. This demonstrates that JMTFA mitigates conflicting between two modalities and improves the collaborative attacking performance in VLP transformers.

% UAPs against VLP models, there are no existing benchmarks for
% comparison. To address this, we construct baselines based on prior
% works that focus on sample-specific attacks against VLP models [35],
% along with variants of our proposed method. These can help verify
% the efficacy of each component and demonstrate the superiority of
% our overall algorithm in learning UAPs.

Given the absence of established benchmarks for baseline comparison, we construct baselines based on gradient-based combination attacks
to compare. We incorporate \textit{FGSM+PWW}, \textit{PGD+PWWS}, \textit{FGSM+BERT-Attack}, \textit{PGD+BERT-Attack} by default settings, and \textit{Co-Attack} that is designed to reduce conflicting when adding two-modality adversarial perturbations as baselines. Table~\ref{tab:jftma_baseline} displays the comparative findings for the VQA and VSR tasks, revealing important insights. Using the strongest attack settings by constraints, JMTFA outperforms combined multimodal adversarial attack and Co-Attack, substantiating the reliability of the analysis. Furthermore, JMTFA surpasses baseline attacks, demonstrating its ability to mitigate conflicts between two modalities and improve collaborative attacking performance in VLP transformers. These results underscore the effectiveness of JMTFA in generating adversarial samples and its potential for advancing research in multimodal adversarial attacks and defenses.

% These adversarial examples contribute to the study of the explainable multimodal transformer fusion and provide insights into how modality input corruptions affect performance

JMTFA adversarial interventions at the multimodal level successfully trick VLP transformer classifiers into making incorrect predictions by deceptive manipulation, which is stronger than the single-modality attack due to perturbing the joint embedding space.  Our findings reveal that the vision modality serves as the primary defense against adversarial attacks in selected VLP transformer models, both in text and image attacks. The limitations of JMTFA can be summarized in the following perspectives. In VQA, generated adversarial text examples are difficult for humans to distinguish from original sentences, due to out-of-domain words. There exists a bias towards ``Yes" answers in VQA and ``True" results in VSR tasks While there is no semantic difference in the content. When using BERT-attack for the language modality with $K=1$, it fails to perform word substitutions, resulting in no improvement in attack performance for text inputs. In addition, we focus on fine-tuned downstream task VLP models with similar structures, leaving the adversarial robustness of generative VLP models as an area for future work.

%% file: sec5_ablation.tex
\section{Ablation Study}
\label{sec:Ablation}

\textbf{Image Importance Feature Selection}. Our visual attack in JMTFA can be applied to all network layers. We speculate that this behavior significantly impacts feature importance on model predictions. To investigate this, we compare the performance of different visual understanding levels. Fig.~\ref{fig:ablationstudyvqa} (a) shows attack accuracy results by selecting various image feature layers. We distinguish between first self-attention, middle self-attention ($6th$  attention layer), and last self-attention layers in single-stream architectures, as well as last self-attention, first cross-attention, and last cross-attention layers in dual-stream architectures. The overall attack accuracy has dropped significantly compared to the original performance of victim VLP transformers. LXMERT achieves the highest attack accuracy, {while VLE is the most vulnerable when the final layer features are used to represent an input image. Empirically, this observation confirms that the early layer preserves low-level information related to the ground truth, {while the middle features are not sensitive enough to determine the final output. By contrast, the later layer learns salient semantic features that contribute to classification accuracy. Based on this observation, we select each victim model's final layer representation for image adversarial sample generation.

% \textbf{Image Importance Feature Selection}. Our visual attack in JMTFA can be applied to all network layers. We speculate that this behavior is influential to impact feature importance on model predictions. Hence, we further compare the performance of different visual understanding levels. Figure.~\ref{fig:ablationstudyvqa} (a) shows attack accuracy results by selecting various image feature layers. Feature comparisons distinguish with first self-attention, middle self-attention ($6th$  attention layer) and last self-attention layers in single-stream architectures, in addition to last self-attention, first cross-attention and last cross-attention layers in dual-stream architectures. The overall attack accuracy has dropped significantly compared to the all original performance of selective VLP transformers. LXMERT achieves the highest attack accuracy, and VLE is the most vulnerable when the final layer features are used to represent an input image. Empirically, this observation further confirms that the early layer reserves the low-level pattern information to the ground truth, and the middle feature are not sensitive enough to determine the final output. By contrast, the later layer learns salient semantic features to denote the classification accuracy. Based on this observation, we select each victim model's final layer representation for image adversarial sample generation.  

\begin{figure}[t]
\centering
\includegraphics[width=\linewidth]{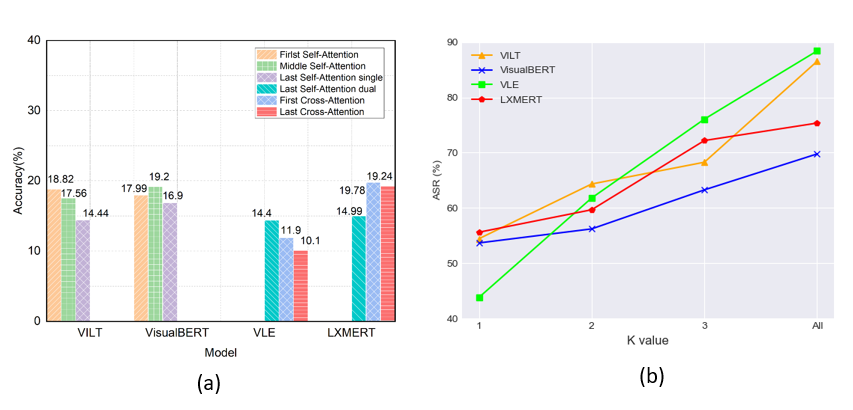}
\caption{(a) Effect of layer selections on the accuracy of vision modality attacks in JMTFA, while maintaining disturbance of all keywords in language modality attacks. (b) Language modality attack performance in JMTFA, measured by ASR, using the last feature output states.}
\label{fig:ablationstudyvqa}
\end{figure}

% 分开的两个图
% \begin{figure}[htbp]
% \centering
% \includegraphics[width=1\linewidth]{figures/vqa_features.png}
% % \includegraphics[width=9cm,height=5.5cm]{figures/}
% \caption[]{} 
% \label{fig:vqafeatures}
% \caption{xxx}
% \end{figure}

% \begin{figure}[htbp]
% \centering
% \includegraphics[width=0.9\linewidth]{figures/vqa_tokens.png}
% % \includegraphics[width=9cm,height=5.5cm]{figures/}
% \caption[]{} 
% \label{fig:vqatokens}
% \caption{xxx}
% \end{figure}

\noindent\textbf{Word Substitution Amount Ranking}. We further analyze the influence of selecting $K$ candidate word choices in textual attacks. First, we consider the worst-case scenario of replacing all words in the sentence except for non-contributing words. Second, we keep a small ratio of representative words $K\leq 3$ while filtering out antonyms. Fig.~\ref{fig:ablationstudyvqa} (b) qualitatively depicts the ASR clearly decreasing as the candidate pool expands. This indicates that certain words significantly impact the prediction, while lower-ranking words have little effect on the outcome. To maintain semantic similarity, we set a threshold of 0.5 for the cosine similarity between the benign and the adversarial questions. This ensures that representative words remain meaningfully and semantically consistent with the original questions. In addition, we avoid using a fixed candidate number, as it can be inflexible in practice. Our experiment ensures that the most plausible perturbations are considered without a strict threshold. In the subsequent human review, a drawback of the BERT language model in predicting candidate words is that it sometimes occasionally results in unnatural sentences.

\begin{figure}[t]
\small
\centering
\includegraphics[width=0.9\linewidth]{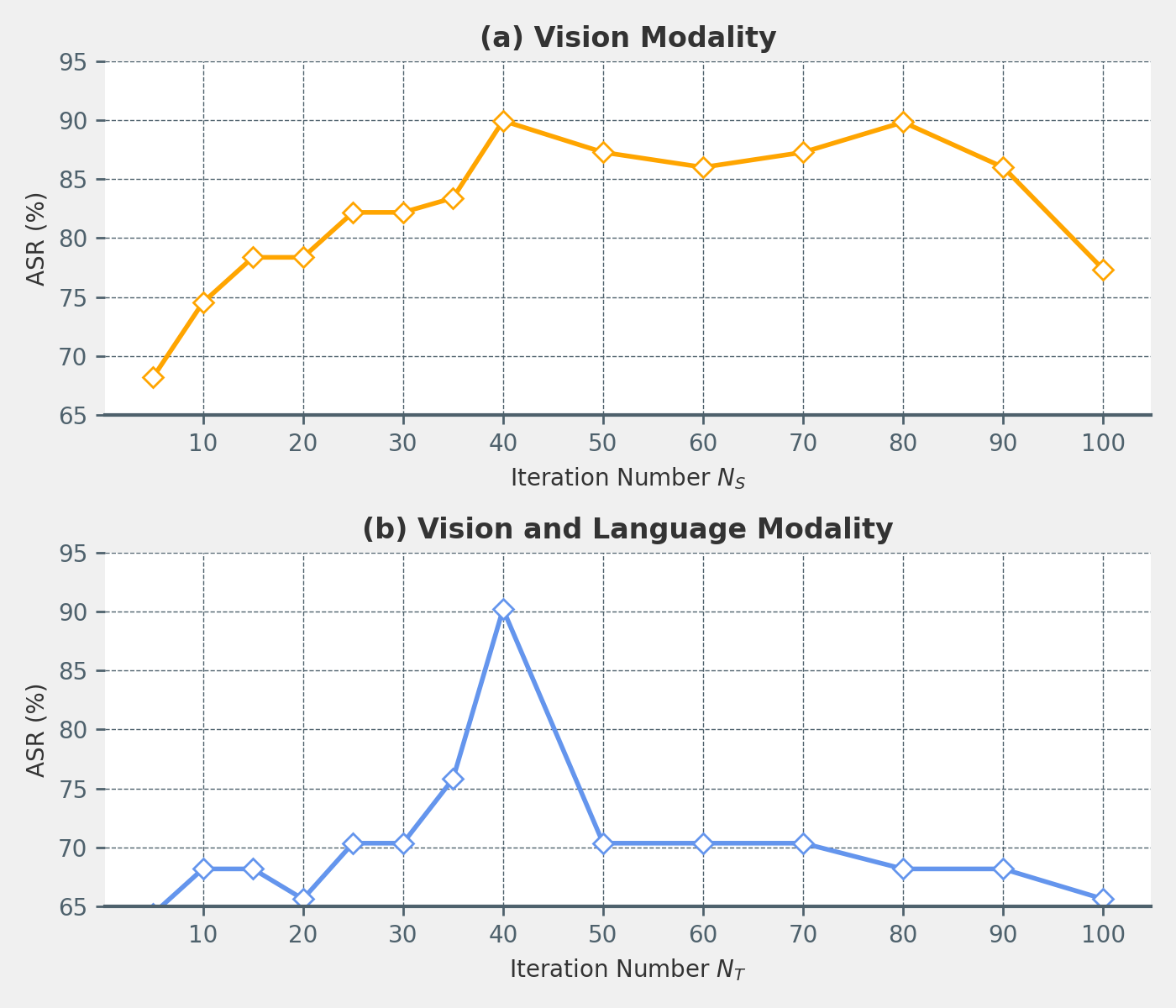}
\caption[]{Investigation of the amount of iterations $N_s$ and $N_t$.}
\label{fig:ablationstudyvsr} 
\end{figure}

\noindent\textbf{Parameter Iteration Analysis}. We investigate how varying the iteration numbers $N_s$ and $N_t$ affect JMTFA. All experiments utilize the VSR dataset and the ViLT model. $N_s$ and $N_t$ represent iterations of single vision modality and two-modality attacks, respectively. The total number of iterations $N_s$ and $ N_t$ ranges from 5 to 100.  As illustrated in Fig.~\ref{fig:ablationstudyvsr} (a), the ASR significantly improves when increasing the number of steps from 5 to 40, reaching its highest performance at $N = 89.94\%$. Subsequently, we examine the two-modality JMTFA iteration in Fig.~\ref{fig:ablationstudyvsr} (b). The ASR peaks at $90.20\%$ and then gradually decreases as $N_s$ increases. We find that a smaller initial iteration number $N_t$ leads to more text perturbations. To balance attack performance with the naturalness of the generated two-modality adversarial examples, we fix the overall iteration number $N$ at 40. This selection provides a balance between effectiveness and maintaining the quality of the adversarial samples.

%% file: sec6_conclusion.tex
\section{Conclusion and Future Work}
\label{sec:Conclusion}

In this paper, we introduce JMTFA, a novel approach that uses important features to execute visual and linguistic adversarial attacks on VLP transformers. This research gap raises important considerations regarding the impact of attacking different modalities on various VLP transformers. Extensive experiments demonstrate that the performance of the adversarial attacks with various settings and analyze key factors to enhance the effectiveness of attacks by two-modality. Our findings contribute to the understanding of vulnerabilities VLP transformers and offer valuable insights for developing more robust VLP models, highlighting the importance of considering both visual and linguistic aspects when designing defense mechanisms for multimodal AI systems. While our study focuses specifically on VLP transformer architectures, the underlying principles of our approach may have broader applicability adapting to across more different multimodal architectures, as the concept of extracting feature saliency tokens or patches. In the future, we seek to investigate the adversarial robustness of Multimodal Large Language Models against sophisticated attacks from malicious users.

%% file: main.bbl
\begin{thebibliography}{10}
\providecommand{\url}[1]{#1}
\csname url@samestyle\endcsname
\providecommand{\newblock}{\relax}
\providecommand{\bibinfo}[2]{#2}
\providecommand{\BIBentrySTDinterwordspacing}{\spaceskip=0pt\relax}
\providecommand{\BIBentryALTinterwordstretchfactor}{4}
\providecommand{\BIBentryALTinterwordspacing}{\spaceskip=\fontdimen2\font plus
\BIBentryALTinterwordstretchfactor\fontdimen3\font minus \fontdimen4\font\relax}
\providecommand{\BIBforeignlanguage}[2]{{%
\expandafter\ifx\csname l@#1\endcsname\relax
\typeout{** WARNING: IEEEtran.bst: No hyphenation pattern has been}%
\typeout{** loaded for the language `#1'. Using the pattern for}%
\typeout{** the default language instead.}%
\else
\language=\csname l@#1\endcsname
\fi
#2}}
\providecommand{\BIBdecl}{\relax}
\BIBdecl

\bibitem{vlp2022du}
Y.~Du, Z.~Liu, J.~Li, and W.~X. Zhao, ``A survey of vision-language pre-trained models,'' in \emph{Proceedings of the International Joint Conference on Artificial Intelligence}, 2022, pp. 5436--5443.

\bibitem{chen2023vlp}
F.-L. Chen, D.-Z. Zhang, M.-L. Han, X.-Y. Chen, J.~Shi, S.~Xu, and B.~Xu, ``Vlp: A survey on vision-language pre-training,'' \emph{Machine Intelligence Research}, vol.~20, no.~1, pp. 38--56, 2023.

\bibitem{lu2019vilbert}
J.~Lu, D.~Batra, D.~Parikh, and S.~Lee, ``Vilbert: pretraining task-agnostic visiolinguistic representations for vision-and-language tasks,'' in \emph{Proceedings of the Conference on Advances Neural Information Processing Systems}, 2019, pp. 13--23.

\bibitem{li2020oscar}
X.~Li, X.~Yin, C.~Li, P.~Zhang, X.~Hu, L.~Zhang, L.~Wang, H.~Hu, L.~Dong, F.~Wei \emph{et~al.}, ``Oscar: Object-semantics aligned pre-training for vision-language tasks,'' in \emph{Proceedings of the European Conference on Computer Vision}.\hskip 1em plus 0.5em minus 0.4em\relax Springer, 2020, pp. 121--137.

\bibitem{zhang2021vinvl}
P.~Zhang, X.~Li, X.~Hu, J.~Yang, L.~Zhang, L.~Wang, Y.~Choi, and J.~Gao, ``Vinvl: Revisiting visual representations in vision-language models,'' in \emph{Proceedings of the IEEE/CVF Conference on Computer Vision and Pattern Recognition}, June 2021, pp. 5579--5588.

\bibitem{antol2015vqa}
S.~Antol, A.~Agrawal, J.~Lu, M.~Mitchell, D.~Batra, C.~Lawrence~Zitnick, and D.~Parikh, ``Vqa: Visual question answering,'' in \emph{Proceedings of the IEEE International Conference on Computer Vision}, 2015, pp. 2425--2433.

\bibitem{wu2017visual}
Q.~Wu, D.~Teney, P.~Wang, C.~Shen, A.~Dick, and A.~Van Den~Hengel, ``Visual question answering: A survey of methods and datasets,'' \emph{Computer Vision and Image Understanding}, vol. 163, pp. 21--40, 2017.

\bibitem{zellers2019recognition}
R.~Zellers, Y.~Bisk, A.~Farhadi, and Y.~Choi, ``From recognition to cognition: Visual commonsense reasoning,'' in \emph{Proceedings of the IEEE/CVF conference on computer vision and pattern recognition}, 2019, pp. 6720--6731.

\bibitem{chen2015microsoft}
X.~Chen, H.~Fang, T.-Y. Lin, R.~Vedantam, S.~Gupta, P.~Doll{\'a}r, and C.~L. Zitnick, ``Microsoft coco captions: Data collection and evaluation server,'' \emph{arXiv preprint arXiv:1504.00325}, 2015.

\bibitem{shayegani2023survey}
E.~Shayegani, M.~A.~A. Mamun, Y.~Fu, P.~Zaree, Y.~Dong, and N.~Abu-Ghazaleh, ``Survey of vulnerabilities in large language models revealed by adversarial attacks,'' \emph{arXiv preprint arXiv:2310.10844}, 2023.

\bibitem{zhang2022towards}
J.~Zhang, Q.~Yi, and J.~Sang, ``Towards adversarial attack on vision-language pre-training models,'' in \emph{Proceedings of the 30th ACM International Conference on Multimedia}, 2022, pp. 5005--5013.

\bibitem{kim2021vilt}
W.~Kim, B.~Son, and I.~Kim, ``Vilt: Vision-and-language transformer without convolution or region supervision,'' in \emph{Proceedings of the International Conference on Machine Learning}.\hskip 1em plus 0.5em minus 0.4em\relax PMLR, 2021, pp. 5583--5594.

\bibitem{li2020visualbert}
L.~H. Li, M.~Yatskar, D.~Yin, C.-J. Hsieh, and K.-W. Chang, ``What does bert with vision look at?'' in \emph{Proceedings of the Association for Computational Linguistics}, 2020, pp. 5265--5275.

\bibitem{iflytek2023vle}
iFLYTEK and HFL, ``Vle: Vision-language encoder,'' \url{https://github.com/iflytek/VLE}, 2023.

\bibitem{tan2019lxmert}
H.~Tan and M.~Bansal, ``Lxmert: Learning cross-modality encoder representations from transformers,'' in \emph{Proceedings of the Empirical Methods in Natural Language Processing and the International Joint Conference on Natural Language Processing}, 2019, pp. 5100--5111.

\bibitem{li2022blip}
J.~Li, D.~Li, C.~Xiong, and S.~Hoi, ``Blip: Bootstrapping language-image pre-training for unified vision-language understanding and generation,'' in \emph{Proceedings of the International conference on machine learning}.\hskip 1em plus 0.5em minus 0.4em\relax PMLR, 2022, pp. 12\,888--12\,900.

\bibitem{dou2022empirical}
Z.-Y. Dou, Y.~Xu, Z.~Gan, J.~Wang, S.~Wang, L.~Wang, C.~Zhu, P.~Zhang, L.~Yuan, N.~Peng \emph{et~al.}, ``An empirical study of training end-to-end vision-and-language transformers,'' in \emph{Proceedings of the IEEE/CVF Conference on Computer Vision and Pattern Recognition}, 2022, pp. 18\,166--18\,176.

\bibitem{szegedy2014intriguing}
C.~Szegedy, W.~Zaremba, I.~Sutskever, J.~Bruna, D.~Erhan, I.~Goodfellow, and R.~Fergus, ``Intriguing properties of neural networks,'' in \emph{Proceedings of the International Conference on Learning Representations}, 2014.

\bibitem{goodfellow2015explaining}
I.~J. Goodfellow, J.~Shlens, and C.~Szegedy, ``Explaining and harnessing adversarial examples,'' in \emph{Proceedings of the International Conference on Learning Representations}, 2015.

\bibitem{madry2018towards}
A.~Madry, A.~Makelov, L.~Schmidt, D.~Tsipras, and A.~Vladu, ``Towards deep learning models resistant to adversarial attacks,'' in \emph{Proceedings of the International Conference on Learning Representations}, 2018.

\bibitem{papernot2016limitations}
N.~Papernot, P.~McDaniel, S.~Jha, M.~Fredrikson, Z.~B. Celik, and A.~Swami, ``The limitations of deep learning in adversarial settings,'' in \emph{Proceedings of the IEEE European Symposium on Security and Privacy}, 2016, pp. 372--387.

\bibitem{su2019one}
J.~Su, D.~V. Vargas, and K.~Sakurai, ``One pixel attack for fooling deep neural networks,'' \emph{IEEE Transactions on Evolutionary Computation}, vol.~23, no.~5, pp. 828--841, 2019.

\bibitem{moosavi2017universal}
S.-M. Moosavi-Dezfooli, A.~Fawzi, O.~Fawzi, and P.~Frossard, ``Universal adversarial perturbations,'' in \emph{Proceedings of the IEEE Conference on Computer Vision and Pattern Recognition}, 2017, pp. 1765--1773.

\bibitem{carlini2017towards}
N.~Carlini and D.~Wagner, ``Towards evaluating the robustness of neural networks,'' in \emph{Proceedings of the IEEE Symposium on Security and Privacy}, 2017, pp. 39--57.

\bibitem{wang2021feature}
Z.~Wang, H.~Guo, Z.~Zhang, W.~Liu, Z.~Qin, and K.~Ren, ``Feature importance-aware transferable adversarial attacks,'' in \emph{Proceedings of the IEEE/CVF international conference on computer vision}, 2021, pp. 7639--7648.

\bibitem{zhang2022enhancing}
Y.~Zhang, Y.-a. Tan, T.~Chen, X.~Liu, Q.~Zhang, and Y.~Li, ``Enhancing the transferability of adversarial examples with random patch.'' in \emph{Proceedings of the International Joint Conference on Artificial Intelligence}, 2022, pp. 1672--1678.

\bibitem{ganeshan2019fda}
A.~Ganeshan, V.~BS, and R.~V. Babu, ``Fda: Feature disruptive attack,'' in \emph{Proceedings of the IEEE/CVF International Conference on Computer Vision}, 2019, pp. 8069--8079.

\bibitem{gao2018black}
J.~Gao, J.~Lanchantin, M.~L. Soffa, and Y.~Qi, ``Black-box generation of adversarial text sequences to evade deep learning classifiers,'' in \emph{Proceedings of the IEEE Security and Privacy Workshops}.\hskip 1em plus 0.5em minus 0.4em\relax IEEE, 2018, pp. 50--56.

\bibitem{eger2019text}
S.~Eger, G.~G. {\c S}ahin, A.~R{\"u}ckl{\'e}, J.-U. Lee, C.~Schulz, M.~Mesgar, K.~Swarnkar, E.~Simpson, and I.~Gurevych, ``Text processing like humans do: Visually attacking and shielding nlp systems,'' in \emph{Proceedings of the North American Chapter of the Association for Computational Linguistics}, 2019, pp. 1634--1647.

\bibitem{boucher2022bad}
N.~Boucher, I.~Shumailov, R.~Anderson, and N.~Papernot, ``Bad characters: Imperceptible nlp attacks,'' in \emph{Proceedings of the IEEE Symposium on Security and Privacy}.\hskip 1em plus 0.5em minus 0.4em\relax IEEE, 2022, pp. 1987--2004.

\bibitem{ren2019generating}
S.~Ren, Y.~Deng, K.~He, and W.~Che, ``Generating natural language adversarial examples through probability weighted word saliency,'' in \emph{Proceedings of the Association for Computational Linguistics}, 2019, pp. 1085--1097.

\bibitem{jin2020bert}
D.~Jin, Z.~Jin, J.~T. Zhou, and P.~Szolovits, ``Is bert really robust? a strong baseline for natural language attack on text classification and entailment,'' in \emph{Proceedings of the Assocaition for artificial intelligence}, vol.~34, no.~05, 2020, pp. 8018--8025.

\bibitem{garg2020bae}
S.~Garg and G.~Ramakrishnan, ``Bae: Bert-based adversarial examples for text classification,'' in \emph{Proceedings of the Conference on Empirical Methods in Natural Language Processing}, 2020, pp. 6174--6181.

\bibitem{iyyer2018adversarial}
M.~Iyyer, J.~Wieting, K.~Gimpel, and L.~Zettlemoyer, ``Adversarial example generation with syntactically controlled paraphrase networks,'' in \emph{Proceedings of the North American Chapter of the Association for Computational Linguistics: Human Language Technologies}, 2018, pp. 1875--1885.

\bibitem{ribeiro2018semantically}
M.~T. Ribeiro, S.~Singh, and C.~Guestrin, ``Semantically equivalent adversarial rules for debugging nlp models,'' in \emph{Proceedings of the Association for Computational Linguistics}, 2018, pp. 856--865.

\bibitem{naik2018stress}
A.~Naik, A.~Ravichander, N.~Sadeh, C.~Rose, and G.~Neubig, ``Stress test evaluation for natural language inference,'' in \emph{Proceedings of the International Conference on Computational Linguistics}, 2018, pp. 2340--2353.

\bibitem{yin2022vlattack}
Z.~Yin, M.~Ye, T.~Zhang, T.~Du, J.~Zhu, H.~Liu, J.~Chen, T.~Wang, and F.~Ma, ``Vlattack: Multimodal adversarial attacks on vision-language tasks via pre-trained models,'' in \emph{Proceedings of the Conference on Advances Neural Information Processing Systems}, 2023.

\bibitem{lu2023set}
D.~Lu, Z.~Wang, T.~Wang, W.~Guan, H.~Gao, and F.~Zheng, ``Set-level guidance attack: Boosting adversarial transferability of vision-language pre-training models,'' in \emph{Proceedings of the IEEE/CVF International Conference on Computer Vision}, 2023, pp. 102--111.

\bibitem{he2023sa}
B.~He, X.~Jia, S.~Liang, T.~Lou, Y.~Liu, and X.~Cao, ``Sa-attack: Improving adversarial transferability of vision-language pre-training models via self-augmentation,'' \emph{arXiv preprint arXiv:2312.04913}, 2023.

\bibitem{han2023ot}
D.~Han, X.~Jia, Y.~Bai, J.~Gu, Y.~Liu, and X.~Cao, ``Ot-attack: Enhancing adversarial transferability of vision-language models via optimal transport optimization,'' \emph{arXiv preprint arXiv:2312.04403}, 2023.

\bibitem{chefer2021generic}
H.~Chefer, S.~Gur, and L.~Wolf, ``Generic attention-model explainability for interpreting bi-modal and encoder-decoder transformers,'' in \emph{Proceedings of the IEEE/CVF International Conference on Computer Vision}, 2021, pp. 397--406.

\bibitem{bach2015pixel}
S.~Bach, A.~Binder, G.~Montavon, F.~Klauschen, K.-R. M{\"u}ller, and W.~Samek, ``On pixel-wise explanations for non-linear classifier decisions by layer-wise relevance propagation,'' \emph{PloS one}, vol.~10, no.~7, p. e0130140, 2015.

\bibitem{abnar2020quantifying}
S.~Abnar and W.~Zuidema, ``Quantifying attention flow in transformers,'' in \emph{Proceedings of the Association for Computational Linguistics}, 2020, pp. 4190--4197.

\bibitem{li2020bert}
L.~Li, R.~Ma, Q.~Guo, X.~Xue, and X.~Qiu, ``Bert-attack: Adversarial attack against bert using bert,'' in \emph{Proceedings of the Conference on Empirical Methods in Natural Language Processing}, 2020, pp. 6193--6202.

\bibitem{zhang2023towards}
J.~Zhang, Y.-C. Huang, W.~Wu, and M.~R. Lyu, ``Towards semantics-and domain-aware adversarial attacks,'' in \emph{Proceedings of the International Joint Conference on Artificial Intelligence}, 2023, pp. 536--544.

\bibitem{dosovitskiy2021an}
A.~Dosovitskiy, L.~Beyer, A.~Kolesnikov, D.~Weissenborn, X.~Zhai, T.~Unterthiner, M.~Dehghani, M.~Minderer, G.~Heigold, S.~Gelly, J.~Uszkoreit, and N.~Houlsby, ``An image is worth 16x16 words: Transformers for image recognition at scale,'' in \emph{Proceedings of the International Conference on Learning Representations}, 2021.

\bibitem{ren2016faster}
S.~Ren, K.~He, R.~Girshick, and J.~Sun, ``Faster r-cnn: Towards real-time object detection with region proposal networks,'' \emph{IEEE transactions on pattern analysis and machine intelligence}, vol.~39, no.~6, pp. 1137--1149, 2016.

\bibitem{kenton2019bert}
J.~D. M.-W.~C. Kenton and L.~K. Toutanova, ``Bert: Pre-training of deep bidirectional transformers for language understanding,'' in \emph{Proceedings of North American Association for Computational Linguistics- Human Language Technologies}, 2019, pp. 4171--4186.

\bibitem{he2022debertav3}
P.~He, J.~Gao, and W.~Chen, ``Debertav3: Improving deberta using electra-style pre-training with gradient-disentangled embedding sharing,'' in \emph{Proceedings of the International Conference on Learning Representations}, 2022.

\bibitem{liu2022vsr}
F.~Liu, G.~E.~T. Emerson, and N.~Collier, ``Visual spatial reasoning,'' \emph{Transactions of the Association for Computational Linguistics}, 2023.

\bibitem{lin2014microsoft}
T.-Y. Lin, M.~Maire, S.~Belongie, J.~Hays, P.~Perona, D.~Ramanan, P.~Doll{\'a}r, and C.~L. Zitnick, ``Microsoft coco: Common objects in context,'' in \emph{Proceedings of the European Conference on Computer Vision}.\hskip 1em plus 0.5em minus 0.4em\relax Springer, 2014, pp. 740--755.

\bibitem{reimers2019sentence}
N.~Reimers and I.~Gurevych, ``Sentence-bert: Sentence embeddings using siamese bert-networks,'' in \emph{Proceedings of the Empirical Methods in Natural Language Processing and the International Joint Conference on Natural Language Processing}, 2019, pp. 3982--3992.

\bibitem{marchi2021cross}
C.~K. Marchi~Fagundes, K.~Stock, and L.~S. Delazari, ``A cross-linguistic study of spatial location descriptions in new zealand english and brazilian portuguese natural language,'' \emph{Transactions in GIS}, vol.~25, no.~6, pp. 3159--3187, 2021.

\end{thebibliography}
